%% file: main.tex
\documentclass[journal]{IEEEtran}

\usepackage{graphicx}
\graphicspath{{fig/}}

\usepackage{microtype}
\usepackage{url}
\usepackage{xcolor}
\definecolor{myorange}{HTML}{FF7F0D}
\definecolor{myblue}{HTML}{1E77B4}
\definecolor{myred}{HTML}{D62728}

\newcommand{\system}[1]{{\small \textsc{#1}}}
\newcommand{\tablesystem}[1]{\textsc{#1}}
\usepackage[cmex10]{amsmath}
\usepackage{amssymb}

\renewcommand{\vec}[1]{\mathbf{#1}}

\usepackage{booktabs}
\usepackage{tabularx}
\usepackage{multirow}
\usepackage{siunitx}
\newcommand{\mytable}{
    \centering
    \renewcommand{\arraystretch}{1.2}
    }
\newcolumntype{C}{>{\centering\arraybackslash}X}
\newcolumntype{L}{>{\raggedright\arraybackslash}X}

\usepackage{url}

\usepackage{cite}  
\usepackage[prependcaption,textsize=scriptsize]{todonotes}
\definecolor{mycolor}{HTML}{FF6600}

\begin{document}

\title{Improved acoustic word embeddings for zero-resource languages using multilingual transfer}
\author{Herman~Kamper \qquad Yevgen Matusevych \qquad Sharon Goldwater
\thanks{H.\ Kamper is with the Department of E\&E Engineering, Stellenbosch University, South Africa (email: kamperh@sun.ac.za).}%
\thanks{Y.\ Matusevych and S.\ Goldwater are with the School of Informatics, University of Edinburgh, UK (email: ymatusev@ed.ac.uk and sgwater@inf.ed.ac.uk).}%
\thanks{We thank C.\ Jacobs for helpful input.
This work is based on research supported in part by the National Research Foundation of South Africa (grant number:\ 120409), a James S.\ McDonnell Foundation Scholar Award (220020374), an ESRC-SBE award (ES/R006660/1), and a Google Faculty Award for~HK.}
}

\markboth{Preprint, 2020}
{Kamper, Matusevych, Goldwater}

\maketitle

\input{abstract}

\IEEEpeerreviewmaketitle

\input{introduction}

\input{related_work}

\input{models}

\input{experiments}
\input{analysis}
\input{conclusion}

\bibliography{mybib}

\end{document}

%% file: abstract.tex
\begin{abstract}
Acoustic word embeddings are fixed-dimensional representations of variable-length speech segments. Such embeddings can form the basis for speech search, indexing and discovery systems when conventional speech recognition is not possible. In \textit{zero-resource} settings where unlabelled speech is the only available resource, we need a method that gives robust embeddings on an arbitrary language. Here we explore multilingual transfer: we train a single supervised embedding model on labelled data from multiple well-resourced languages and then apply it to unseen zero-resource languages. We consider three multilingual recurrent neural network (RNN) models: a classifier trained on the joint vocabularies of all training languages; a Siamese RNN trained to discriminate between same and different words from multiple languages; and a correspondence autoencoder (CAE) RNN trained to reconstruct word pairs. In a word discrimination task on six target languages, all of these models outperform state-of-the-art unsupervised models trained on the zero-resource languages themselves, giving relative improvements of more than 30\% in average precision. When using only a few training languages, the multilingual CAE performs better, but with more training languages the other multilingual models perform similarly. Using more training languages is generally beneficial, but improvements are marginal on some languages. We present probing experiments which show that the CAE encodes more phonetic, word duration, language identity and speaker information than the other multilingual models.
\end{abstract}


%% file: introduction.tex
\section{Introduction}

The dependence of automatic speech recognition~(ASR) systems on transcribed speech data remains a major hurdle for developing speech technology in new languages.
Researchers in \textit{zero-resource speech processing} aim to develop 
unsupervised methods that can learn directly from unlabelled speech audio~\cite{jansen+etal_icassp13,versteegh+etal_sltu16,dunbar+etal_asru17}.
Several tasks have been tackled.
In unsupervised term discovery (UTD), the goal is to find recurring word- or phrase-like patterns in an unlabelled speech collection~\cite{park+glass_taslp08}.
In query-by-example search, the aim is to identify utterances containing instances of a given spoken query~\cite{hazen+etal_asru09,zhang+glass_asru09,levin+etal_icassp15}.
Full-coverage segmentation and clustering aims to tokenise an entire speech set into word-like units~\cite{elsner+shain_emnlp17,lee+etal_tacl15,rasanen+etal_interspeech15,kamper+etal_asru17}.

In these applications, speech segments of different durations need to be compared.
One approach is to use dynamic time warping (DTW), but this is computationally expensive and can be inaccurate~\cite{rabiner+etal_tassp78}.
Levin et al.~\cite{levin+etal_asru13} therefore proposed an alignment-free approach: a speech segment of arbitrary length is embedded in a fixed-dimensional space.
The resulting vectors are referred to as \textit{acoustic word embeddings}.
Segments can then be compared by simply calculating a distance between their vectors in the embedding space.
This requires a mapping such that instances of the same word type (lexical item) have similar embeddings.
Several supervised and unsupervised acoustic word embedding methods have since been proposed~\cite{chung+etal_interspeech16,kamper+etal_icassp16,settle+etal_interspeech17,audhkhasi+etal_stsp17,kamper_icassp19}.

While unsupervised methods are useful in that they can be used in zero-resource settings, there is still a large performance gap compared to supervised methods~\cite{levin+etal_asru13,kamper_icassp19}.
Here we investigate whether supervised modelling can still be used to obtain accurate embeddings on a language for which no labelled data is available.
Specifically, we propose to exploit labelled resources from languages where these are available, allowing us to take advantage of state-of-the-art supervised modelling methods, but to then apply the resulting model to zero-resource languages for which no labelled data is available.

For this transfer learning approach, we
consider three multilingual acoustic embedding models.
A classifier recurrent neural network (RNN) is trained on the joint vocabularies of several well-resourced languages~\cite{settle+livescu_slt16}.
A Siamese RNN is trained with a contrastive loss on multiple languages so that word instances of the same type have similar embeddings while others are pushed away~\cite{settle+livescu_slt16}.
Finally, the correspondence autoencoder RNN (\system{CAE-RNN}) uses an encoder-decoder structure to reconstruct one word given another word of the same type as input~\cite{kamper+etal_icassp19}.
We use six well-resourced languages from the GlobalPhone corpus~\cite{schultz+etal_icassp13} for training, and evaluate the models using a word discrimination task on six target languages which we treat as zero-resource.

Our goal is to find the best acoustic embedding approach for zero-resource languages.
To show that multilingual transfer is the best in this setting, we need to compare it to an appropriate unsupervised monolingual baseline.
We extend~\cite{kamper_icassp19}, where an unsupervised \system{CAE-RNN} was trained on discovered terms from a UTD system, and use the same approach to train unsupervised \system{ClassifierRNN} and \system{SiameseRNN} models on unlabelled data from zero-resource languages, similar to~\cite{huang+etal_arxiv18}.
For the first time, these three unsupervised models are compared.
The \system{CAE-RNN} performs best.
To answer our main research question, we compare this unsupervised approach to the supervised multilingual \system{CAE-RNN}, \system{ClassifierRNN} and \system{SiameseRNN}.
This is an extension of~\cite{kamper+etal_icassp20}, where the first two were considered.
The \system{SiameseRNN}---a popular method in the monolingual supervised setting~\cite{settle+livescu_slt16}---has not been previously used for multilingual transfer.
Across the six zero-resource languages, all three multilingual models outperform unsupervised \system{CAE-RNNs} trained on the respective evaluation languages.

Of the three approaches, the multilingual \system{CAE-RNN} performs better when only a few well-resourced training languages are used, but with more languages there is little difference between the models.
Here we also introduce a new variant of the multilingual \system{CAE-RNN} which conditions its decoder on the training language identity.
This gives small but consistent improvements on the zero-resource languages.
Finally, inspired by the analysis of supervised monolingual acoustic word embedding models presented in ~\cite{matusevych+etal_baics20}, we perform probing experiments to see how the multilingual models organise their embedding spaces.
We find that, compared to the other multilingual models, the \system{CAE-RNN} captures more phonetic, word duration and speaker information.
Source code is released at {\tt \small \url{https://github.com/kamperh/globalphone_awe}}.

%% file: related_work.tex
\section{Related Work}

Acoustic word embeddings are fixed-dimensional vector representations of arbitrary-length spoken words~\cite{levin+etal_asru13}.
The goal is to map different instances of the same spoken word to similar embeddings, while mapping words of different types to embeddings that are far apart.
This allows variable-length speech segments to be efficiently compared directly in a fixed-dimensional embedding space without any alignment (which can be slow).
Acoustic embeddings are therefore being used 
increasingly in downstream tasks such as 
full-coverage segmentation and clustering~\cite{kamper+etal_asru17}, query-by-example speech search~\cite{levin+etal_icassp15,wang+etal_icassp18,yuan+etal_interspeech18}, and spoken content retrieval~\cite{audhkhasi+etal_stsp17,chen+etal_slt18}.

Our goal is to learn an acoustic word embedding model that can be applied in a zero-resource setting where no labelled data is available in the target language.
Unsupervised modelling can be used directly in this setting.
A simple but effective early unsupervised approach is \textit{downsampling}~\cite{levin+etal_asru13,holzenberger+etal_interspeech18}: a fixed number of equally spaced frames are used to represent a speech segment.
More recent work has turned to neural methods.
Unsupervised auto-encoding recurrent neural networks (RNNs) use an encoder-decoder structure to try to reconstruct variable-duration input~\cite{chung+etal_interspeech16,audhkhasi+etal_stsp17,holzenberger+etal_interspeech18}.
Instead of trying to reconstruct the input exactly, the correspondence autoencoder RNN (\system{CAE-RNN})~\cite{kamper_icassp19} is trained to reconstruct another speech segment predicted to be of the same type as the input.
Since labelled data is not available, 
a UTD system---itself unsupervised---is used to find word-like pairs predicted to be of the same unknown type.
This model outperformed downsampling and an \system{AE-RNN} in~\cite{kamper_icassp19}.
However, it still falls far short from supervised models trained on labelled data.
Here we investigate whether supervised modelling
can still be used to obtain
embeddings on a language for which no labelled data is available.

Early supervised acoustic word embedding methods relied on a reference vector approach: the DTW distances between an input segment and a fixed reference set are calculated, followed by
supervised dimensionality reduction~\cite{levin+etal_asru13}.
This has since been outperformed by supervised neural approaches relying on convolutional~\cite{kamper+etal_icassp16,haque+etal_icassp19} and RNN-based~\cite{settle+livescu_slt16,settle+etal_interspeech17} architectures.
These networks can be trained with a classification loss on labelled isolated words, where features from an intermediate layer before the final softmax are used as embeddings, or with a contrastive loss, where a Siamese network with tied branches explicitly pushes embeddings of words of the same type together while pushing different words apart.
Most studies have reported that the contrastive loss performs better, although improvements over the classifier are sometimes small~\cite{kamper+etal_icassp16,settle+livescu_slt16}.
More recent work has started to incorporate surrounding context~\cite{yuan+etal_interspeech18,shi+etal_arxiv19,palaskar+etal_icassp19}, in some cases specifically to capture semantic relationships~\cite{chen+etal_slt18,chung+glass_interspeech18}.
Some models explicitly embed written and spoken words  together~\cite{bengio+heigold_interspeech14,ghannay+etal_evalnlp16,he+etal_iclr17,audhkhasi+etal_stsp17,settle+etal_icassp19,jung+etal_asru19}.
Although these directions are important, we focus on the original question of learning embedding models that map together spoken words of the same type.

In all of the above studies,
models were always subsequently applied to the \textit{same language} as the one used during training.
One approach we consider here is to train a supervised model on (multiple) well-resourced languages and then apply it to a zero-resource language.
Our aim is to determine the best embedding approach when confronted with a zero-resource setting: is it better to apply a (potentially multilingual) supervised model on an unseen language, or is it better to train in an unsupervised fashion on the target language itself?
Apart from studies explicitly looking at acoustic embeddings, this question also relates to several other lines of research.

Most importantly, our work is inspired by studies showing the benefit of using multilingual bottleneck features as frame-level representations for zero-resource languages~\cite{vesely2012language,yuan+etal_icassp17,hermann+etal_arxiv18,menon+etal_interspeech19}:
a
frame-level network is trained jointly on several well-resourced languages (normally to predict context-dependent triphone HMM states) and is then applied to an unseen language.
In~\cite{ondel+etal_arxiv19}, multilingual data was also used for discovering acoustic units.
As in these studies, our findings show the advantage of learning from labelled data in well-resourced languages when processing an unseen low-resource language---here at the word rather than subword level.
We are also inspired by studies in ASR aiming to build a single system that can transcribe speech from a number of different input languages.
Early approaches considered joint training of a single set of HMM-GMM models on multiple languages~\cite{schultz+waibel_speechcom01,niesler_speechcom07}, while more recent work has turned to end-to-end neural networks~\cite{toshniwal+etal_icassp18,tong+etal_speechcom18,cho+etal_slt18,adams+etal_acl19}.
While these models are also typically trained jointly on multiple languages (as we do here), they are then applied to test data from languages on which they are trained. In contrast, the languages to which our models are applied are never seen during training.

Training a machine learning model on one task and then applying it to another related problem is becoming an attractive way to leverage existing data or models---a methodology referred to as \textit{transfer learning}~\cite{pan+yang_kde09}.
Our approach can be described as multilingual transfer since we
train a single embedding model on labelled data from several well-resourced languages and then embed  speech from a target zero-resource language.
While in many transfer learning settings there is a subsequent fine-tuning step with some (limited) labelled data from the target domain (referred to as \textit{inductive transfer learning}), we directly apply a multilingual model to a new language (called \textit{transductive transfer learning} in~\cite{pan+yang_kde09,ruder_phd19}).

%% file: models.tex
\section{Acoustic Word Embedding Models}
\label{sec:models}

For obtaining acoustic word embeddings on a zero-resource language, we compare unsupervised models trained on within-language unlabelled data to supervised models trained on pooled labelled data from multiple well-resourced languages.
We consider three different recurrent neural network (RNN) architectures.
All three uses gated recurrent units~\cite{chung+etal_arxiv14}.
Below we first describe how each architecture is used for supervised multilingual modelling, and then explain how the same architectures are used for unsupervised monolingual modelling.

\subsection{Supervised multilingual acoustic word embeddings}
\label{sec:supervised_models}

Given labelled data from several well-resourced languages, we consider three supervised multilingual acoustic embedding models.
We use $X = \vec{x}_1, \vec{x}_2, \ldots, \vec{x}_T$ to denote a sequence of frame-level acoustic feature vectors, with each vector $\vec{x}_t \in \mathbb{R}^D$, e.g.\ $D$-dimensional MFCCs.

\subsubsection{Classifier RNN}
\label{sec:supervised_classifier_rnn}

\begin{figure}[!b]
    \centering
    \includegraphics[scale=0.85]{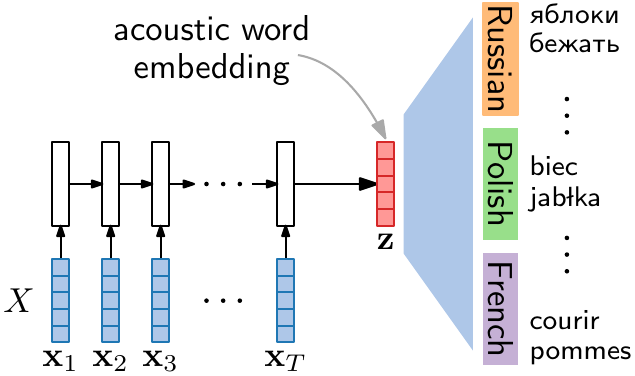}
    \caption{The multilingual \tablesystem{ClassifierRNN} is trained jointly on all the training languages to classify which word type an input segment $X$ belongs to. Our model is trained on data from six languages (three shown here as illustration).
    }
    \label{fig:classifier_rnn}
\end{figure}

Given a true isolated word segment $X$ from any of the training languages, the \system{ClassifierRNN} predicts the word type of that segment.
Formally, it is trained using the multiclass log loss, given as $J(X) = - \sum_{k = 1}^K y_k \, \log f_k(X)$ for a single training example, where $K$ is the size of the joint vocabulary over all the training languages, $y_k \in \{0, 1\}$ is an indicator for whether $X$ is an instance of word type $k$, and $\boldsymbol{f}(X) \in [0, 1]^K$ 
is the predicted distribution over the joint vocabularies of all the languages.
As shown in Figure~\ref{fig:classifier_rnn}, an input sequence $X$ is presented to an encoder RNN shared between all training languages.
A fixed-dimensional acoustic word embedding $\vec{z} \in \mathbb{R}^M$ is then obtained by transforming the final hidden state of the RNN~\cite{audhkhasi+etal_stsp17}.
This embedding is fed into a softmax layer to produce $\boldsymbol{f}(X)$.
Embeddings can therefore be obtained for speech segments from a language not seen during training.

\subsubsection{Siamese RNN}
\label{sec:supervised_siamese_rnn}

\begin{figure}[!b]
    \centering
    \includegraphics[scale=0.85]{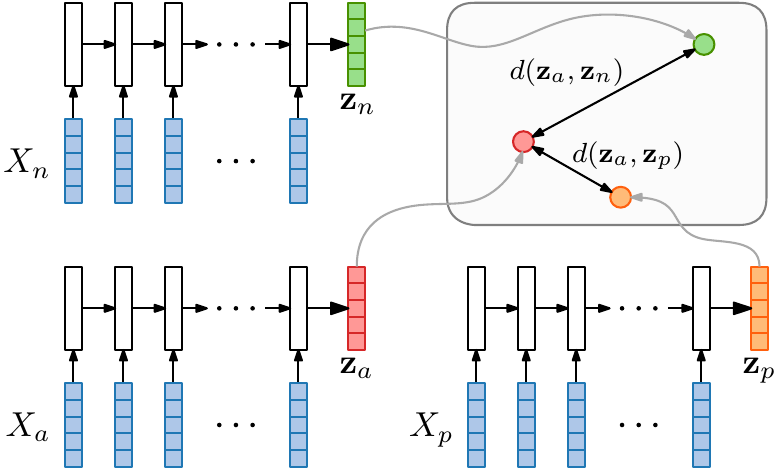}
    \caption{The multilingual \tablesystem{SiameseRNN} is trained jointly on all the training languages so that the distance between the embedding of an anchor word $\mathbf{z}_a$  and a positive example $\mathbf{z}_p$ is smaller (by some margin) than the distance between the anchor and a negative example $\mathbf{z}_n$.
    The three RNNs shown in the figure share the same set of parameters.}
    \label{fig:siamese_rnn}
\end{figure}

In the \system{ClassifierRNN} we obtain embeddings from an intermediate layer with the hope that instances of the same word type will have similar embeddings.
Rather than doing this implicitly through a classification loss, the \system{SiameseRNN} explicitly optimises a similarity loss between embeddings~\cite{bromley+etal_ijpr93,settle+livescu_slt16}.
Formally, input sequences $X_a$, $X_p$, $X_n$ are each passed through an RNN to produce embeddings $\vec{z}_a$, $\vec{z}_p$, $\vec{z}_n$, as illustrated in Figure~\ref{fig:siamese_rnn}.
$X_a$ and $X_p$ are instances of the same word type (subscripts indicate \textit{anchor} word and \textit{positive} word) while $X_n$ is a different word (\textit{negative}).
The RNN's parameters are optimised using the contrastive loss~\cite{weinberger+etal_jmlr09,chechik+etal_jmlr10}:
$J(X_a, X_p, X_n) = \max \left\{ 0, m + d(\vec{z}_a , \vec{z}_p ) - d(\vec{z}_a , \vec{z}_n ) \right\}$,
with $d(\cdot)$ the squared Euclidean distance  and $m$ a margin parameter.
This loss is at a minimum when all embedding pairs $(\vec{z}_a, \vec{z}_p)$ of the same word type are more similar
by a margin $m$ than pairs $(\vec{z}_a, \vec{z}_n)$ of different types.
To sample negative items we use the online semi-hard mining scheme~\cite{schroff+etal_cvpr15}: for each positive pair in a mini-batch the most difficult negative item is used satisfying the constraint $d(\vec{z}_a, \vec{z}_d) < d(\vec{z}_a, \vec{z}_n)$, except if there is no such item in which case the negative example with the largest distance is used.
The multilingual \system{SiameseRNN} is trained jointly on all the well-resourced training languages.

\subsubsection{Correspondence autoencoder RNN}
\label{sec:multilingual_cae_rnn}

\begin{figure}[!b]
    \centering
    \includegraphics[scale=0.85]{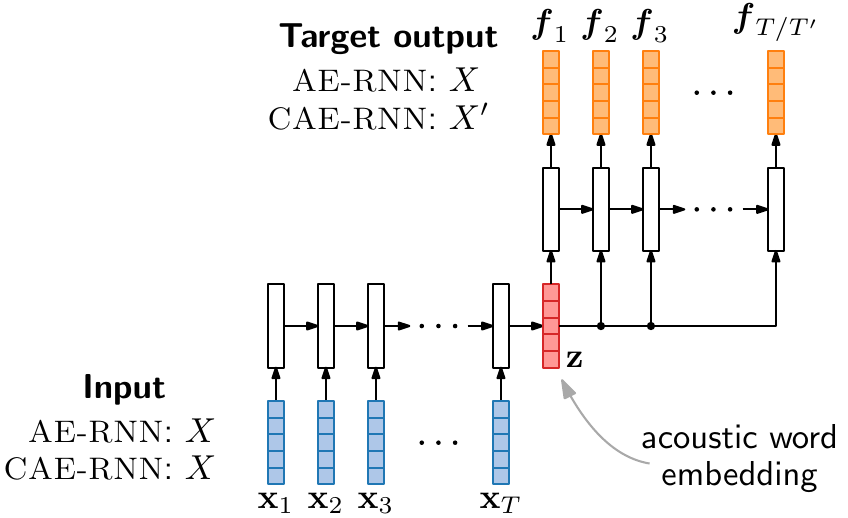}
    \caption{The \tablesystem{AE-RNN} is trained to reconstruct its input $X$ (a speech segment) from the latent acoustic word embedding $\vec{z}$.
    The \tablesystem{CAE-RNN} is trained to reconstruct one segment $X'$ when presented with another segment $X$ as input.
    For the supervised multilingual \tablesystem{CAE-RNN} (\S\ref{sec:multilingual_cae_rnn}), true word pairs $(X, X')$ are used from the well-resourced training languages.
    For the unsupervised monolingual \tablesystem{CAE-RNN (UTD)} (\S\ref{sec:unsup_cae_rnn}), an unsupervised term discovery system is used to find word pairs $(X, X')$.}
    \label{fig:cae_rnn}
\end{figure}

The \system{CAE-RNN} was originally developed as an unsupervised monolingual embedding method~\cite{kamper_icassp19}, as explained below. 
But given true word segments from forced alignments, it can also be trained in a supervised way.
Here we train a single \system{CAE-RNN} by pooling word segments from several well-resourced training languages.
Formally, the \system{CAE-RNN} is trained on pairs of speech segments $(X, X')$, with 
$X = \vec{x}_1, \ldots, \vec{x}_{T}$ and $X' = \vec{x}'_1, \ldots, \vec{x}'_{T'}$
containing different instances of the same word type.
As illustrated in Figure~\ref{fig:cae_rnn},
$X$ is fed into an encoder RNN, which produces the acoustic word embedding $\vec{z}$. 
This embedding is used to condition a decoder RNN which attempts to reconstruct $X'$.
The loss for a single training pair is therefore $J(X, X') = \sum_{t = 1}^{T'} ||\vec{x}'_t - \boldsymbol{f}_t(X) ||^2$, where $\boldsymbol{f}_t(X)$ is the $t^{\textrm{th}}$ decoder output conditioned on the embedding $\mathbf{z}$.

The idea is that \system{CAE-RNN} embeddings should be invariant to properties not common to the two segments (e.g. speaker, channel), while capturing aspects that are (e.g. word identity).
Here we hope that this type of invariance can be learned from well-resourced languages and then transferred to an unseen language.
This motivation for the \system{CAE-RNN} is similar to that of the \system{SiameseRNN}.
But the former relies on a softer reconstruction loss while the latter uses an explicitly contrastive loss.
Also, the \system{CAE-RNN} is trained only on positive pairs, while the \system{SiameseRNN} incorporates negative items.

\subsection{Language conditioning}
\label{sec:models_language_conditioning}

As a contribution not considered in prior work, we present model variants which use training language identity information in the decoder of a model.
The idea is that this would allow a model to capture more language-specific properties after the embedding layer, while common aspects across languages are captured in the shared encoder.
This could produce embeddings that generalise better to unseen languages.

Instead of having only a single shared softmax layer after the embedding layer in the multilingual \system{ClassifierRNN} (\S\ref{sec:supervised_classifier_rnn}), we consider a model where we add separate fully connected layers for each training language, each language-specific branch terminating in its own softmax output layer.
For the multilingual \system{CAE-RNN} (\S\ref{sec:multilingual_cae_rnn}), together with the acoustic embedding $\mathbf{z}$, we additionally append a language embedding at each decoder time-step.
The language embedding  matrix  is  updated  jointly  with  the  rest  of  the network.

\subsection{Unsupervised monolingual acoustic word embeddings}
\label{sec:unsupervised_models}

An alternative to the transfer learning approaches in \S\ref{sec:supervised_models} for obtaining acoustic embeddings on a zero-resource language is to train an unsupervised embedding model directly on unlabelled data in the target language.
Four unsupervised monolingual embedding models are described below.
The unsupervised \system{ClassifierRNN} and \system{SiameseRNN} have not been considered before.
The experiment in \S\ref{sec:exp_unsupervised}
is therefore also the first comparison of these unsupervised approaches.

\subsubsection{Unsupervised autoencoder and correspondence autoencoder RNNs}
\label{sec:unsup_cae_rnn}

The unsupervised autoencoding RNN (\system{AE-RNN}) of~\cite{chung+etal_interspeech16} is trained on unlabelled speech segments to reproduce its input.
Formally, given an input speech segment $X$, the loss for a single training example is 
$J(X) = \sum_{t = 1}^T \left|\left|\vec{x}_t - \boldsymbol{f}_t(X)\right|\right|^2$,
with $\boldsymbol{f}_t(X)$ the $t^\textrm{th}$ decoder output, 
as also illustrated in Figure~\ref{fig:cae_rnn}. 

We next consider the unsupervised variant of the \system{CAE-RNN}, first proposed in~\cite{kamper_icassp19}.
In contrast to the supervised multilingual \system{CAE-RNN} (\S\ref{sec:multilingual_cae_rnn}), in the unsupervised setting we do not have access to transcriptions from which to construct input-output training pairs.
We therefore apply 
an
unsupervised term discovery~(UTD) system~\cite{park+glass_taslp08,jansen+vandurme_asru11} 
to an unlabelled speech collection in the target zero-resource language, discovering pairs of word segments predicted to be of the same unknown type.
These are used as input-output pairs $(X, X')$ to the \system{CAE-RNN}, as shown in Figure~\ref{fig:cae_rnn}. 
We denote the resulting unsupervised monolingual model as \system{CAE-RNN (UTD)}.
In all cases, we first pretrain a \system{CAE-RNN} using the AE loss above before switching to the CAE~loss.

\subsubsection{Unsupervised classifier and Siamese RNNs}

This strategy of using a UTD system to discover training targets for the \system{CAE-RNN (UTD)} can also be employed with the other architectures of \S\ref{sec:supervised_models}.
Instead of predicting the word type of an input segment (\S\ref{sec:supervised_classifier_rnn}), the monolingual unsupervised \system{ClassifierRNN (UTD)} is trained to predict the cluster labels from the UTD system.
The unsupervised \system{SiameseRNN (UTD)} uses matching pairs from the UTD system as positive examples, while sampling negative items randomly from UTD terms with different cluster labels.
These two unsupervised models have not been considered in any previous work.

%% file: experiments.tex
\section{Experimental Setup}

\subsection{Data}

We perform all
our experiments on the GlobalPhone corpus of read speech~\cite{schultz+etal_icassp13}.
We treat six languages as 
our target
zero-resource
languages: 
Spanish (ES), Hausa (HA), Croatian (HR), Swedish (SV), Turkish (TR) and Mandarin~(ZH).
Each language has on average 16 hours of training, 2 hours of development and 2 hours of test data.
For training supervised multilingual embedding models, six
other GlobalPhone languages are chosen as well-resourced languages:
Czech (CS), French (FR), Polish (PL), Portuguese~(PT), Russian (RU) and Thai (TH).\footnote{In~\cite{kamper+etal_icassp20}, Bulgarian was also used, but we decided to exclude it here because of poor-quality forced alignments (determined through manual inspection).}
Each well-resourced language has on average 21 hours of labelled training data.

\subsection{Model training, architectures and implementation}

We pool the data from the well-resourced languages and train three supervised multilingual models (\S\ref{sec:supervised_models}).
The supervised multilingual \system{ClassifierRNN} is trained jointly on true word segments from the six training languages, obtained from forced alignments.
The number of word types per language is limited to 10k, giving a total of 70k output classes (more classes did not give improvements).
Rather than considering all possible word pairs from all languages when training the multilingual \system{CAE-RNN} and \system{SiameseRNN} models, we sample 300k true word pairs from the combined data. Using more pairs did not improve development performance, but increased training time. 

Since we do not use
transcriptions for the unsupervised monolingual embedding models (\S\ref{sec:unsupervised_models}), we apply the UTD system of~\cite{jansen+vandurme_asru11} to each of the training sets of the six zero-resource languages.\footnote{As described at {\scriptsize \tt \url{https://github.com/eginhard/cae-utd-utils}}.}
Roughly 36k predicted word pairs are extracted in each language.
The unsupervised monolingual \system{AE-RNN}, \system{CAE-RNN}, \system{ClassifierRNN} and \system{SiameseRNN} models are all trained on the same set of UTD terms.
For the \system{AE-RNN}, an alternative would have been to use segments randomly sampled from the speech audio, but UTD-discovered segments gave slightly better performance in~\cite{kamper_icassp19}.

Since development data would not be available in a zero-resource setting, we perform development experiments on labelled data from yet another language: German (DE).
We used this data to tune the vocabulary size for the \system{ClassifierRNN}, the number of pairs for the \system{CAE-RNN} and \system{SiameseRNN}models, the margin for the  \system{SiameseRNN}, the language embedding size of the multilingual \system{CAE-RNN}, and the number of training epochs.
Other hyperparameters are set as in~\cite{kamper_icassp19}.

All models are implemented in TensorFlow. 
Speech audio is parametrised as $D = \text{13}$ dimensional static Mel-frequency cepstral coefficients (MFCCs).
We use an embedding dimensionality of $M = \text{130}$ throughout, since downstream systems such as the segmentation and clustering system of~\cite{kamper+etal_asru17} are
constrained to embedding sizes of this order.
All encoder-decoder models have three encoder and three decoder
unidirectional RNN layers, each with 400 units.
The same encoder structure is used for the \system{ClassifierRNN} and \system{SiameseRNN}.
The \system{SiameseRNN} uses a margin of $m = 0.25$.
Pairs are presented to the \system{CAE-RNN} models in both input-output directions.
For the multilingual \system{CAE-RNN} using language conditioning (\S\ref{sec:models_language_conditioning}), a language embedding with 200 dimensions is used.
Models are trained using Adam optimisation~\cite{kingma+ba_iclr15} with a learning rate of 0.001.

\subsection{Evaluation and baselines}

We want to measure the intrinsic quality of the resulting acoustic word embeddings without being tied to a particular downstream
system architecture.
We therefore use a word discrimination task designed for this purpose~\cite{carlin+etal_icassp11}.
In the \textit{same-different} task,
we are given a pair of acoustic segments, each a true word,
and we must decide whether the segments are examples of
the same or different words.
To evaluate a particular embedding method, a set of isolated test words are first embedded.
For every word pair in this set, the cosine distance
between their embeddings is calculated. Two words can then
be classified as being of the same or different type based on
some distance threshold, and a precision-recall curve is obtained by
varying the threshold. The area under this curve is used as final
evaluation metric, referred to as the average precision (AP).
We are particularly interested in obtaining embeddings that are speaker invariant.
As in ~\cite{hermann+etal_csl20}, we therefore calculate AP by only taking the recall over instances of the same word spoken by different speakers (i.e., we consider the more difficult setting since a model does not get credit for recalling the same word if it is said by the same speaker).

As an additional unsupervised baseline embedding method, we use downsampling~\cite{levin+etal_asru13,holzenberger+etal_interspeech18}
by keeping 10 equally-spaced MFCC vectors from a segment
with appropriate interpolation, giving a 130-dimensional embedding.
Finally, we report same-different performance when using
DTW alignment cost
to predict whether word segments are the same or not.
For this baseline, we additionally include delta and double-delta MFCCs (which was found to be beneficial).

\section{Experimental Results}

Our main goal is to find the best acoustic word embedding approach for
an unseen zero-resource language.
We hypothesise that multilingual acoustic word embedding models trained on labelled data from well-resourced languages will be superior to monolingual unsupervised embedding models trained directly on unlabelled data from the target zero-resource language.
To test this hypothesis, we first need to establish
the best unsupervised embedding approach to serve as an appropriate baseline.
We start, therefore, in \S\ref{sec:exp_unsupervised} by comparing different monolingual unsupervised models.
In section \S\ref{sec:negative}, we briefly consider development experiments to find the best multilingual model variants, specifically looking at the effect of language conditioning (\S\ref{sec:models_language_conditioning}).
We then answer our main research question in \S\ref{sec:exp_unsupervised_vs_multilingual} by comparing the best unsupervised approach to the best multilingual embedding approaches.

\subsection{What is the best acoustic word embedding model in the purely unsupervised case?}
\label{sec:exp_unsupervised}

\begin{table}[!b]
    \mytable
    \caption{AP (\%) on development data for the zero-resource languages using unsupervised acoustic word embedding models.
    For reference, results from supervised model variants (making use of ground truth word segments and labels) are also given.
    The best results of unsupervised and supervised models are highlighted.}
    \begin{tabularx}{1.0\linewidth}{@{}L c S[table-format=2.1] S[table-format=2.1] S[table-format=2.1] S[table-format=2.1] S[table-format=2.1]@{}}
        \toprule
        Model & ES & HA & HR & SV & {TR} & ZH \\
        \midrule
        \underline{\textit{Unsupervised models:}} \\[2pt]
        {DTW} & 22.3 & 27.9 & 16.9 & 14.2 & \textbf{22.2} & 19.3 \\
        {Downsampling} & 14.6 & 20.4 & 14.8 & 8.4 & 16.3 & 15.3 \\
        \tablesystem{AE-RNN (UTD)} & 15.3 & 11.2 & 12.6 & 6.9 & 12.7 & 14.3 \\
        \tablesystem{CAE-RNN (UTD)} & \textbf{28.8} & \textbf{36.0} & \textbf{23.6} & \textbf{15.6} & 20.4 & 20.8 \\
        \tablesystem{ClassifierRNN (UTD)} & 15.0 & 15.4 & 13.6 & 7.6 & 9.8 & 15.4 \\
        \tablesystem{SiameseRNN (UTD)} & 21.3 & 5.4 & 6.2 & 8.7 & 11.9 & \textbf{28.4} \\[2pt]
        \underline{\textit{Supervised models:}} \\[2pt]
        \tablesystem{CAE-RNN (GT)} & \textbf{71.9} & \textbf{77.8} & \textbf{76.0} & 59.7 & 68.7 & 84.6 \\
        \tablesystem{ClassifierRNN (GT)} & 68.3 & 70.4 & 75.3 & 50.3 & 61.0 & 83.5\\
        \tablesystem{SiameseRNN (GT)} & 69.3 & 73.3 & 73.1 & \textbf{64.0} & \textbf{73.8} & \textbf{90.0} \\
        \bottomrule
    \end{tabularx}
    \label{tbl:unsupervised}
\end{table}

Table~\ref{tbl:unsupervised} shows the AP on development data for the unsupervised monolingual models from \S\ref{sec:unsupervised_models} applied to the six zero-resource evaluation languages.
All the neural models are trained on UTD pairs.
For reference, the performance of supervised monolingual models trained using forced alignments and ground truth labels is also given. This can be seen as an upper bound where a perfect UTD system is available.

As in~\cite{kamper_icassp19}, we see that the unsupervised \system{CAE-RNN} outperforms downsampling and the \system{AE-RNN} on all six zero-resource languages.
Here we additionally show that it also outperforms \system{ClassifierRNN} and \system{SiameseRNN} models when these are trained on UTD terms (the only exception is the \system{SiameseRNN} on Mandarin achieving a higher AP of 28.4\%).
The \system{CAE-RNN (UTD)} even outperforms DTW on five of the languages, which is noteworthy since DTW has access to the full sequences for discriminating between words.

The supervised model variants (bottom section, Table~\ref{tbl:unsupervised}) still perform better than the unsupervised models (top section).
In the supervised setting, the best performance is achieved by either the \system{CAE-RNN} or \system{SiameseRNN}, depending on the language.
This is different from 
the unsupervised setting where the \system{CAE-RNN} is better than the \system{SiameseRNN} in most cases.
One interpretation could be that, since the \system{SiameseRNN} is trained in a discriminative fashion, it might be more susceptible to erroneous matches by the UTD system; the \system{CAE-RNN}, in contrast, is trained using a softer reconstruction-like loss which could be more robust to UTD errors.

\begin{table}[!t]
    \mytable
    \caption{AP (\%) on Hausa development data when systematically reducing the noise from the UTD discovery procedure.}
    \begin{tabularx}{\linewidth}{@{}L S[table-format=2.1] S[table-format=2.1]@{}}
        \toprule
        Training set & \tablesystem{CAE-RNN} & \tablesystem{SiameseRNN} \\
        \midrule
        1. UTD terms & 36.0 & 5.4 \\ 
        2. UTD with corrected end-pointing & 29.8 & 7.7 \\ 
        3. UTD with corrected pair labels & 44.5 & 37.7 \\ 
        4. UTD with corrected end-pointing and labels & 69.7 & 60.7 \\ 
        \bottomrule
    \end{tabularx}
    \label{tbl:unsupervised_analysis_hausa}
\end{table}

To support this claim, Table~\ref{tbl:unsupervised_analysis_hausa} presents results on Hausa, where we 
systematically reduce the noise from the UTD system.
Using forced alignments, we compare 
each UTD term with the ground truth word token with which it overlaps most.
We first correct end-pointing: the start and end frame positions predicted by the UTD system are changed to match that of the ground truth word with maximal overlap (row 2).
We then keep the UTD end-pointing, but only train on pairs where the ground truth labels match (row 3).
In row 4, we correct both the end-pointing and only train on pairs with the same true label.
We see that as the UTD pairs are updated with correct labels (row 3) and additionally also end-pointing (row 4), the difference in performance between the \system{CAE-RNN} and \system{SiameseRNN} shrinks to around 9\% absolute in AP.

\subsection{The effect of language conditioning}
\label{sec:negative}

In \S\ref{sec:models_language_conditioning} we described model variants of the multilingual \system{ClassifierRNN} and \system{CAE-RNN}  where the decoder of a model has access to language-specific information.
To see if this leads to better generalisation to unseen languages, we compare model variants with and without language conditioning.
In the \system{ClassifierRNN}, we found that split prediction branches give slightly worse performance than using a single shared softmax layer.
In contrast, Table~\ref{tbl:results_conditioning} shows that the multilingual \system{CAE-RNN} improves slightly on the development data of most languages when a language embedding is added at each decoder time-step. (This improvement was also obtained on the German development data.)
In the tables below we also denote the \system{CAE-RNN} with language conditioning as \system{CAE-RNN-LC}.

\begin{table}[!t]
    \mytable
    \caption{AP (\%) on development data when adding language conditioning (LC) in the decoder of the multilingual \tablesystem{CAE-RNN}.}
    \begin{tabularx}{1.0\linewidth}{@{}L c S[table-format=2.1] S[table-format=2.1] S[table-format=2.1] S[table-format=2.1] S[table-format=2.1]@{}}
        \toprule
        Multilingual model & ES & HA & HR & SV & {TR} & ZH \\
        \midrule
        \tablesystem{CAE-RNN} & 50.0 & 53.7 & \textbf{41.3} & 30.1 & 38.9 & 50.8 \\
        \tablesystem{CAE-RNN-LC}
        & \textbf{54.6} & \textbf{57.6} & 40.9 & \textbf{33.6} & \textbf{43.0} & \textbf{55.0} \\
        \bottomrule
    \end{tabularx}
    \label{tbl:results_conditioning}
\end{table}

\subsection{Does transfer from a multilingual model produce better embeddings than unsupervised learning on the target language?}
\label{sec:exp_unsupervised_vs_multilingual}

Based on the above experiments, we use the \system{CAE-RNN (UTD)} as our monolingual unsupervised baseline and we use the variant of the multilingual \system{CAE-RNN} with language conditioning.
We now turn to our main question: is a multilingual acoustic word embedding model trained on labelled but external data (\S\ref{sec:supervised_models}) superior to an unsupervised monolingual model trained on unlabelled in-domain data from the target zero-resource language?

\begin{table}[!t]
    \mytable
    \caption{AP (\%) on test data for the zero-resource languages. The unsupervised \tablesystem{CAE-RNN}s are trained separately for each zero-resource language on segments from a UTD system applied to unlabelled monolingual data.
    The multilingual models are trained on ground truth word segments obtained by pooling labelled training data from six well-resourced languages.
    The best overall result on each language is highlighted.}
    \begin{tabularx}{1.0\linewidth}{@{}L c S[table-format=2.1] S[table-format=2.1] S[table-format=2.1] S[table-format=2.1] S[table-format=2.1]@{}}
        \toprule
        Model & ES & HA & HR & SV & {TR} & ZH \\
        \midrule
        \underline{\textit{Unsupervised models:}} \\[2pt]
        {DTW} & 36.2 & 23.8 & 17.0 & 27.8 & 16.2 & 35.9 \\
        {Downsampling} & 24.0 & 15.9 & 14.2 & 18.9 & 11.0 & 26.6 \\
        \tablesystem{CAE-RNN} (UTD) & 49.7 & 27.8 & 26.5 & 28.7 & 16.0 & 33.3 \\[2pt]
        \underline{\textit{Multilingual models:}} \\[2pt]
        \tablesystem{CAE-RNN-LC} & \textbf{72.8} & 47.3 & 42.9 & \textbf{46.4} & \textbf{35.1} & 51.6 \\
        \tablesystem{ClassifierRNN} & 68.8 & \textbf{47.7} & \textbf{43.5} & 45.1 & 35.0 & \textbf{54.5} \\
        \tablesystem{SiameseRNN} & 65.8 & 43.7 & 39.3 & 44.6 & 28.2 & 46.1 \\
        \bottomrule
    \end{tabularx}
    \label{tbl:results_test}
\end{table}

\begin{figure*}[!t]
    \centering
    \includegraphics[width=\linewidth]{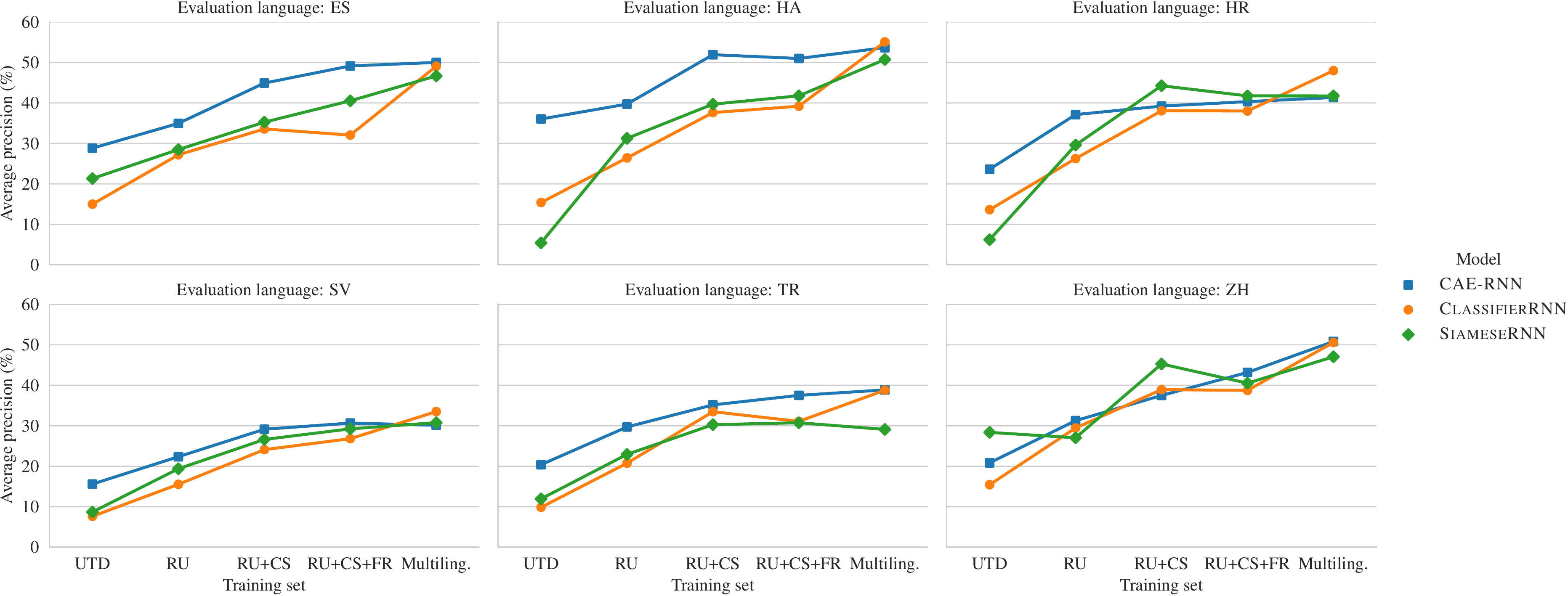}
    \caption{AP (\%) on development data for the six zero-resource language, as multilingual models are trained on one (RU), two (RU+CS), three (RU+CS+FR) and all six (multiling.) training languages. Scores from unsupervised models trained on UTD segments from the evaluation language are also given.}
    \label{fig:adding_languages}
\end{figure*}

Table~\ref{tbl:results_test} shows the performance of monolingual unsupervised and supervised multilingual acoustic word embedding models applied to test data
from the six zero-resource languages.
We see that the multilingual models consistently outperform the unsupervised models across
all six zero-resource languages.
Of the three supervised multilingual models, the \system{SiameseRNN} performs worst.
The relative performance of
the multilingual \system{CAE-RNN} and \system{ClassifierRNN}
models is not consistent over the six zero-resource evaluation
languages, with one model working better on some languages
while another works better on others.
Nevertheless, all three multilingual models consistently outperform the best unsupervised monolingual model
trained directly on the target languages, showing the benefit of
incorporating data from well-resourced languages where labels are available.

Despite the improvements of the multilingual over the unsupervised approach, the multilingual models never outperforms the best monolingual supervised models of Table~\ref{tbl:unsupervised}.\footnote{The two tables here actually report performance on different sets (development and test data, respectively). But this statement is true:
the multilingual models in Table~\ref{tbl:results_test} are always worse than their supervised monolingual counterparts when applied to the same development data.
}
The multilingual models are therefore still not reaching the upper bound from supervised monolingual training.

It is interesting to note that, of the monolingual supervised models (bottom section Table~\ref{tbl:unsupervised}), the \system{SiameseRNN} is one of the best models, outperforming the monolingual supervised \system{CAE-RNN} on three languages.
In contrast, the multilingual \system{SiameseRNN} never performs as well as the multilingual \system{CAE-RNN} when applied to unseen zero-resource languages (Table~\ref{tbl:results_test}).
This could again be due to the discriminative nature of the \system{SiameseRNN}: it is explicitly trained to discriminate between pairs of words from several well-resourced training languages.
Even though it might succeed in this task, this does not necessarily lead to generalisation to unseen languages.
Table~\ref{tbl:seen_unseen_analysis} shows the results when the two models are compared on development data from languages seen during training.
The results show that the multilingual \system{SiameseRNN} outperforms the \system{CAE-RNN} on five out of six of the training languages.
This does not happen even once in Table~\ref{tbl:results_test} when considering performance on unseen languages, which shows that the multilingual \system{SiameseRNN} does not generalise as well.

\begin{table}[!t]
    \mytable
    \caption{AP (\%) achieved by multilingual models when applied to development data from the seen training languages.
    }
    \begin{tabularx}{1.0\linewidth}{@{}L c c c c c c@{}}
        \toprule
        Model & RU & CS & FR & PL & TH & PO \\
        \midrule
        \tablesystem{CAE-RNN-LC} & \textbf{45.6} & 62.6 & 46.9 & 45.3 & 80.8 & 64.7 \\
        \tablesystem{SiameseRNN} &  44.2 & \textbf{76.7} & \textbf{47.8} & \textbf{47.8} & \textbf{85.8} & \textbf{72.7} \\
        \bottomrule
    \end{tabularx}
    \label{tbl:seen_unseen_analysis}
\end{table}

%% file: analysis.tex
\section{Further Analysis}

In this analysis section we look at how the number and choice of training languages impact performance.
We also perform probing experiments to obtain a better understanding of the resulting embedding spaces for the different models.

\subsection{Are more languages beneficial for training multilingual embedding models?}
\label{sec:exp_training_languages}

Figure~\ref{fig:adding_languages} shows development performance on each of the zero-resource language as multilingual models are trained on one (RU), two (RU+CS), three (RU+CS+FR) and all six~(multiling.) well-resourced training languages.
Here we use the multilingual \system{CAE-RNN} without language conditioning (\S\ref{sec:models_language_conditioning}).
For reference, the results from monolingual unsupervised models are also given as the first point in each subplot. We can therefore interpret each plot as giving the results as we increase the number of training languages from zero (UTD) to six.

Results are not entirely consistent across the evaluation languages, but we can identify the following general trends.
Firstly, using labelled data from a single training language (Russian) gives improvements over unsupervised embeddings for all six evaluation languages and all three model variants (except for the Russian \system{SiameseRNN} applied to Mandarin).
Secondly, when using fewer training languages, 
the multilingual \system{CAE-RNN} outperforms the multilingual \system{ClassifierRNN} and \system{SiameseRNN} in most cases.
But, as we increase the number of training languages, all three models improve and it becomes less clear which is superior. On several languages the multilingual \system{ClassifierRNN} gives similar or better performance than the \system{CAE-RNN}.
Although the \system{SiameseRNN} also improves as we add training languages, it is not the top-performing model on any of the six evaluation languages.
Thirdly, the effect of additional training languages differs based on the evaluation language.
Focusing only on the \system{CAE-RNN}, AP improves systematically as we add languages for Spanish, Turkish and Mandarin.
But on Hausa, Croatian and Swedish, performance seems to plateau with two training languages (on Croatian this might not be surprising since both Russian and Czech are from the same language family).

Generally speaking, we can therefore conclude that
adding more training languages does not deteriorate performance, but on some languages improvements are marginal.
In all cases, however, using labelled training data from even a single well-resourced language improves performance over an unsupervised model trained only on the target language.

\subsection{Is the training language choice important?}

\begin{figure}[!t]
    \centering
    \includegraphics[width=0.525\linewidth]{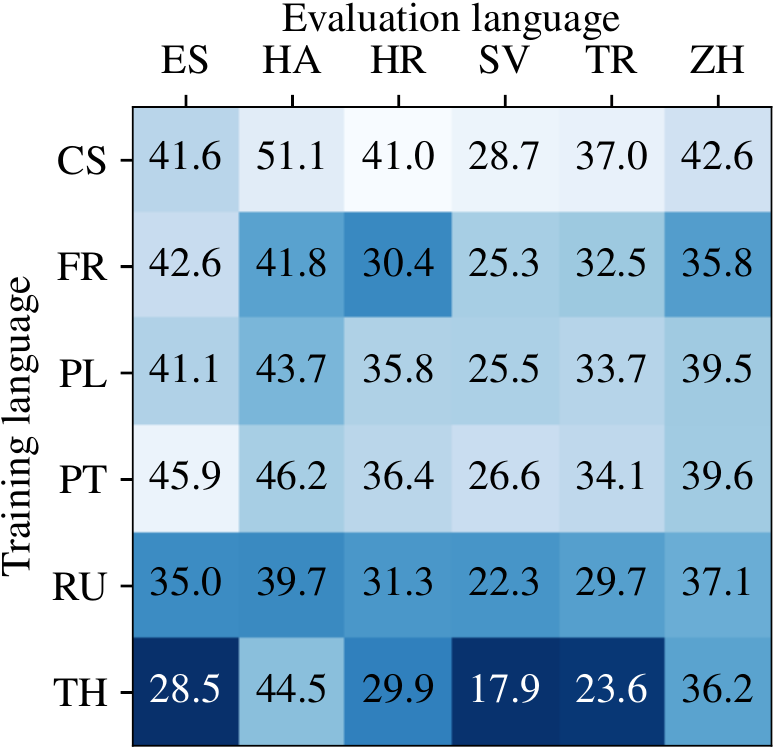}
    \caption{AP (\%) on development data for the six zero-resource languages (columns) when applying different monolingual \tablesystem{CAE-RNN} models, each trained on labelled data from a well-resourced language (rows). Heatmap colours are normalised for each zero-resource language (i.e.\ per column).
    }
    \label{fig:crosslingual}
\end{figure}

To investigate the impact of the particular choice of training language, we train supervised monolingual \system{CAE-RNN}s on each of the well-resourced languages and then apply these
to each of the zero-resource languages.
Results are given in Figure~\ref{fig:crosslingual}.
The choice of well-resourced language can greatly impact performance.
On Spanish, using Portuguese is better than any other language, and on Croatian the  monolingual Czech system performs, showing the benefit of training on languages from the same family.
The supervised Thai model performs worst on all the Indo-European languages but not on Hausa and Mandarin.

Although performance can differ dramatically based on the source-target language pair, all of the systems in Figure~\ref{fig:crosslingual} outperform the unsupervised monolingual
\system{CAE-RNN}s trained on UTD pairs from the target language (Table~\ref{tbl:unsupervised}), except for the Thai-Spanish pair.
Thus, training on just a single well-resourced language (even if it is not in the same language family) is beneficial.
Furthermore, although adding languages does not always lead to improvements in Figure~\ref{fig:adding_languages}, the multilingual \system{CAE-RNN} trained on all six languages (Table~\ref{tbl:results_test}, Figure~\ref{fig:adding_languages}) outperforms all of the supervised monolingual models in Figure~\ref{fig:crosslingual} on the evaluation languages.
The performance effects of training language choice therefore diminish
as more training languages are used.

\subsection{How are the embedding spaces organised?}

In previous sections we compared models using the same-different task.
In some cases there was little to separate the models, e.g., there are only small differences between the supervised multilingual \system{CAE-RNN}, \system{ClassifierRNN} and \system{SiameseRNN} when training on all six languages (right-most points, Figure~\ref{fig:adding_languages}).
Rather than using the same-different task, here we try to directly characterise the organisation
of the models' embedding spaces.
This is valuable, not only for the sake of the analysis itself, but also from a technological perspective: it can help us anticipate potential failure modes when using the embeddings in downstream speech systems. 
As in the analysis of acoustic word embeddings in~\cite{matusevych+etal_baics20}, we use a number of tests to probe the embedding spaces.
The focus in~\cite{matusevych+etal_baics20} was on monolingual models, while here we focus on the multilingual embedding models (\S\ref{sec:supervised_models}).
Instead of reporting the findings from all the tests we performed, we only report those that showed some differences between the multilingual models.

\begin{figure}[!t]
    \centering
    \includegraphics[width=\linewidth]{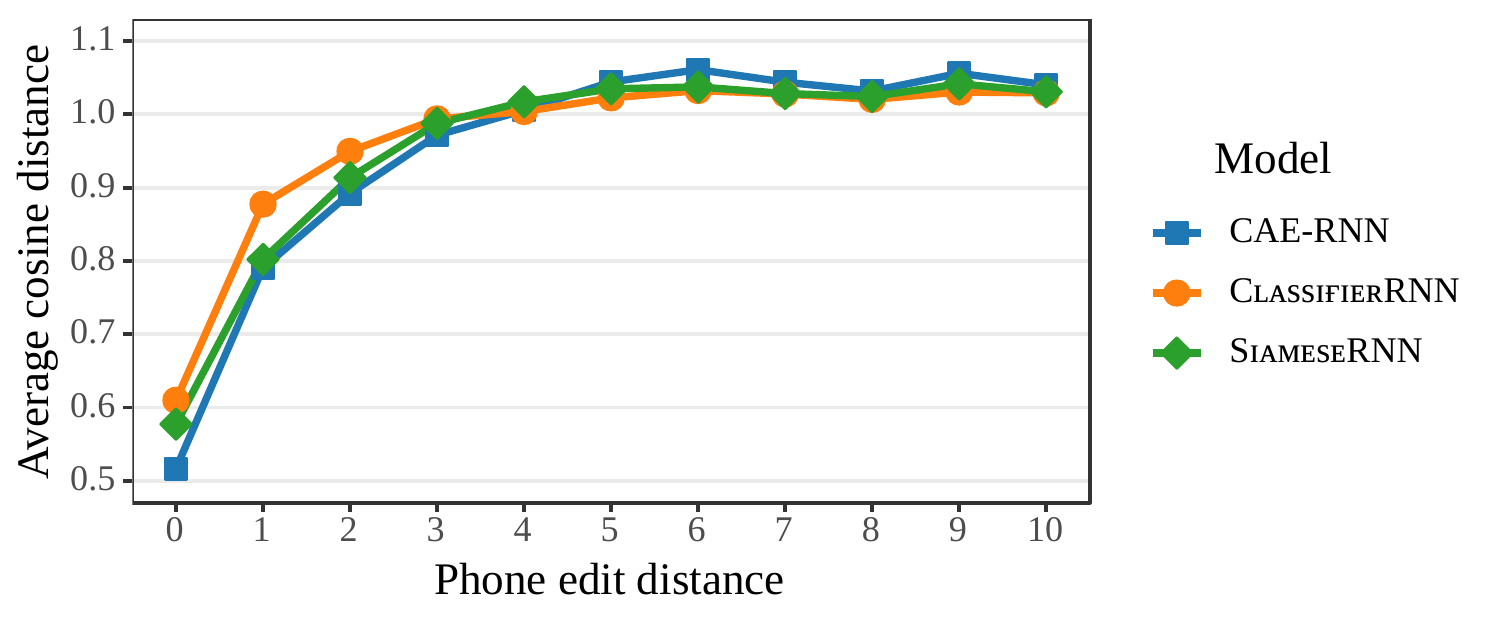}
    \caption{Average cosine distances between pairs of embedded Hausa words as a function of their phone edit distance (development data, multilingual models).}
    \label{fig:cosine-edit-sup}
\end{figure}

We first consider whether the embedding spaces encode the \textit{degree} of similarity between words.
High AP in the same-different task indicates that instances of the same word are closer to each other in an embedding space than different words.
But this does not tell us whether words that are similar but not identical are close to each other.
To measure this, we consider the distance between pairs of embedded words as a function of the phone edit distance between them.
The result for Hausa is shown in Figure~\ref{fig:cosine-edit-sup}.
For all three multilingual embedding models, the average cosine distance between word pairs is lowest when the words are identical (a phone edit distance of $0$), and this distance systematically increases as the words become more phonetically different.
This effect seems strongest for the \system{CAE-RNN}, which has the lowest embedding distance when words are identical, but the largest distance when words differ by $5$ or more edits.
This suggests that embedding spaces are organised according to the words' phonetic similarity, and that this is more so in the multilingual \system{CAE-RNN}.

Next, we consider whether the embedding spaces capture information about word duration and speaker identity.
To do this, we train and test linear classifiers (multi-class logistic regression models) and linear regression models to, respectively, predict the speaker identity of an embedded test word and its duration in milliseconds.
In each case, we train a model on 80\% of the embedded words from the development data of a language and then test the model on the held-out 20\%.

We first consider word duration prediction, looking at the goodness of fit ($R^2$) of the corresponding linear regression models. 
Table~\ref{tbl:dur_regression} shows that, while all three multilingual models perform well on this task, the \system{CAE-RNN} performs best on all six languages. Note that better performance on this task does not necessarily imply better word discrimination performance; this could depend on the language and the specific downstream setting under consideration.
Nevertheless, this analysis shows that the \system{CAE-RNN} seems to capture differences in duration more than the other models (at this salient level measured by linear~regression).

\begin{table}[!t]
    \mytable
    \caption{Goodness of fit ($R^2$, proportion of variance explained) of the regression models predicting absolute word duration on development data for the zero-resource languages.}
    \begin{tabularx}{1.0\linewidth}{@{}L c S[table-format=1.2] S[table-format=1.2] S[table-format=1.2] S[table-format=1.2] S[table-format=1.2] S[table-format=1.2] @{}}
        \toprule
        Model & ES & HA & HR & SV & TR & ZH \\
        \midrule
        \underline{\textit{Unsupervised models:}} \\[2pt]
        {Downsampling}    & 0.47 & 0.72 & 0.57 & 0.36 & 0.51 & 0.69 \\[2pt]
        \underline{\textit{Multilingual models:}} \\[2pt]
        \tablesystem{CAE-RNN}       & \textbf{0.94} & \textbf{0.96} & \textbf{0.96} & \textbf{0.95} & \textbf{0.95} & \textbf{0.94}  \\
        \tablesystem{ClassifierRNN} & 0.79 & 0.76 & 0.81 & 0.78 & 0.71 & 0.71 \\
        \tablesystem{SiameseRNN}    & 0.82 & 0.78 & 0.84 & 0.81 & 0.75 & 0.76 \\
        \bottomrule
    \end{tabularx}
    \label{tbl:dur_regression}
\end{table}        

\begin{table}[!t]
    \mytable
    \caption{Speaker classification accuracy (\%) on development data for the zero-resource languages. The highest scores are highlighted in red (underline) and the lowest in blue (italics).}
    \begin{tabularx}{1.0\linewidth}{@{}L c S[table-format=2.1] S[table-format=2.1] S[table-format=2.1] S[table-format=2.1] S[table-format=2.1] S[table-format=2.1] @{}}
        \toprule
        Model & ES & HA & HR & SV & TR & ZH \\
        \midrule
        \underline{\textit{Unsupervised models:}} \\[2pt]
        {Downsampling} & 31.0 & {\textcolor{myred}{\underline{43.8}}} & {\textcolor{myred}{\underline{38.3}}} & 30.9 & 29.8 & {\textcolor{myred}{\underline{53.0}}} \\[2pt]
        \underline{\textit{Multilingual models:}} \\[2pt]
        \tablesystem{CAE-RNN} & {\textcolor{myred}{\underline{34.3}}} & 38.7 & 33.9 & {\textcolor{myred}{\underline{36.4}}} & {\textcolor{myred}{\underline{36.4}}} & 42.1 \\
        \tablesystem{ClassifierRNN} & 28.3 & 28.5 & 27.9 & {\textcolor{myblue}{\it 24.9}} & 30.4 & 32.0 \\
        \tablesystem{SiameseRNN} & {\textcolor{myblue}{\it 25.7}} & {\textcolor{myblue}{\it 25.4}} & {\textcolor{myblue}{\it 26.4}} & 27.3 & {\textcolor{myblue}{\it 26.4}} & {\textcolor{myblue}{\it 31.0}} \\
        \bottomrule
    \end{tabularx}
    \label{tbl:spk_classif}
\end{table}

We next consider whether the embeddings allow for accurate speaker classification.
Table~\ref{tbl:spk_classif} shows that embeddings from the multilingual \system{CAE-RNN} allow for the best speaker classification results of the three multilingual models.
The \system{SiameseRNN} generally performs the worst.
This indicates that, at a level that can be captured by a linear classifier, the multilingual \system{SiameseRNN}'s embeddings are more speaker-invariant then those of the \system{CAE-RNN}.
This is surprising since the \system{CAE-RNN} generally performs better in cross-speaker word discrimination (Table~\ref{tbl:results_test}, Figure~\ref{fig:adding_languages}).
However, being able to classify speaker using a linear projection of the embeddings does not necessarily imply worse word discrimination performance.
Furthermore, for some downstream tasks it could actually be beneficial to retain some speaker information in the embeddings. 

Finally, we are interested to see whether the multilingual embedding models separate the embedding spaces according to language.
Again, being able to partition the embedding space according to language does not necessarily imply better or worse word discrimination performance, but it can be useful to know whether this happens for settings where words from multiple languages might be embedded at the same time.
Moreover, if the models did perfectly separate the embedding spaces according to training language, the models would presumably not transfer to unseen languages.
For the analysis, we embed at the same time triphones from five languages: RU, CS, FR, DE and ZH.
Three of these languages are seen training languages, while two are unseen.
This allows us to determine if the embeddings encode language identity information better for seen languages.
A set of 25 common triphones are used, e.g., [ana] and [tam]. To ensure that the triphones are genuinely identical across languages, we unified the existing GlobalPhone transcriptions for these languages by converting them into the International Phonetic Alphabet format.
As before, we apply a linear classifier on the models' embeddings.

\begin{figure}[]
    \centering
    \includegraphics[width=1.0\linewidth]{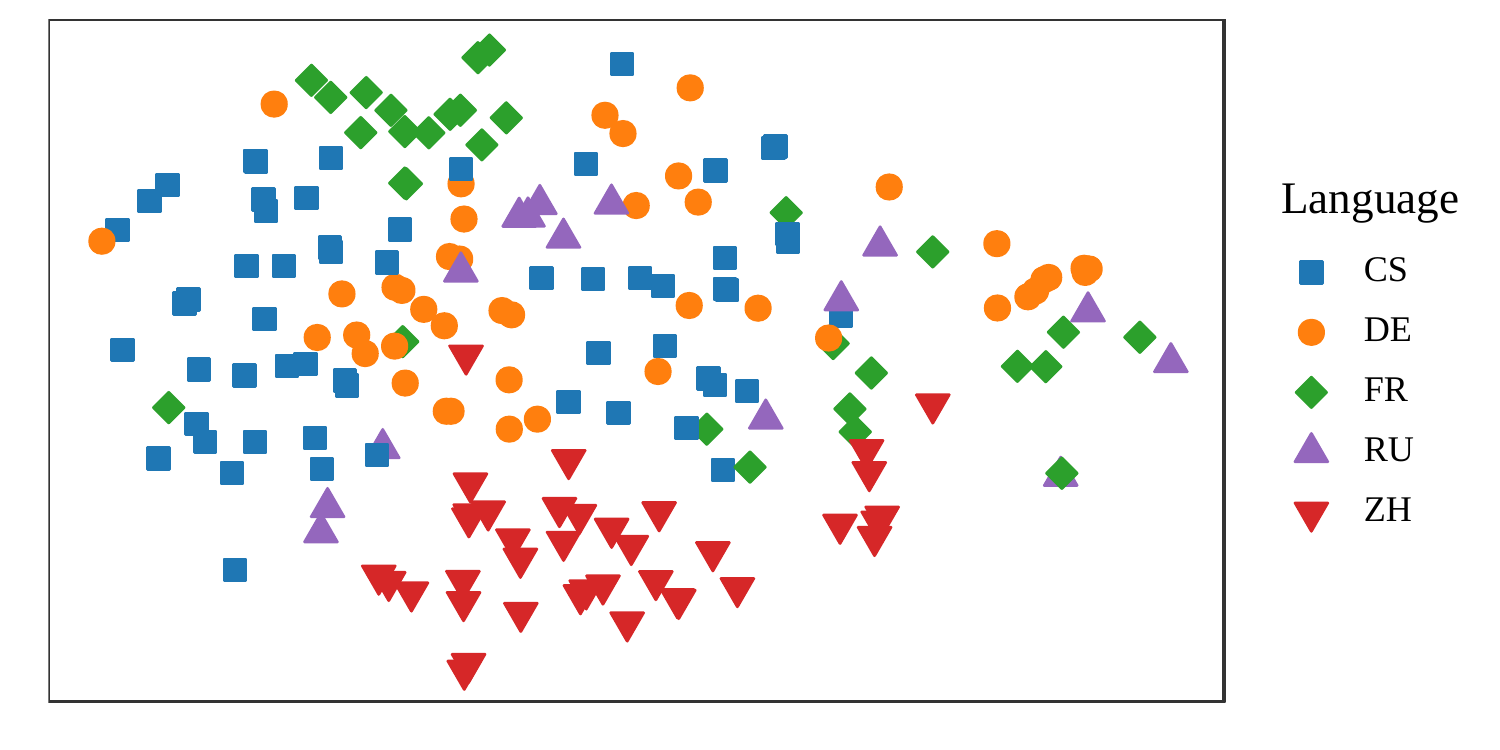}
    \caption{Two-dimensional t-SNE plots of the triphone [tat] in five languages embedded using the multilingual \tablesystem{CAE-RNN}.}
    \label{fig:tSNE}
\end{figure}

Table~\ref{tbl:lang-class} shows that the multilingual \system{CAE-RNN}'s embeddings allows for the best language identification.
Adding more training languages generally improves language classification performance, but this effect is stronger for the \system{ClassifierRNN} and \system{SiameseRNN} than for the \system{CAE-RNN} (which actually performs best when only trained on three languages).
An analysis of the individual $F1$-scores (not shown here) per language shows there are no consistent differences between seen (RU, CS, FR) and unseen (DE, ZH) languages. Figure~\ref{fig:tSNE} shows a visualisation of the embedded instances of the [tat] triphone in all five languages for
the multilingual \system{CAE-RNN}.
While triphones from the same language seem to cluster together, there is a lot of overlap across languages, potentially enabling the model to generalise the acquired phonetic properties of training languages to unseen languages. Note, however, that such generalisation may not work equally well for all languages: in the figure there seems to be more overlap between the training languages (RU, CS, FR) and DE (another Indo-European language) than between the training languages and ZH (a Sino-Tibetan language).

\begin{table}
    \mytable
    \caption{Accuracy (\%) of linear classifiers predicting the language identity of triphones from the development data of five languages (three seen, two unseen).}
    \begin{tabularx}{1.0\linewidth}{@{}L c c c c@{}}
        \toprule
        Model & {RU} & {RU+CS} & {RU+CS+FR} & {Multilingual} \\
        \midrule
        \tablesystem{CAE-RNN}  & \textbf{61.8}  & \textbf{63.7} & \multicolumn{1}{c}{\textbf{66.8}} & \textbf{65.2} \\
        \tablesystem{ClassifierRNN} & 54.3 & 54.2 & 56.2 & 61.8 \\
        \tablesystem{SiameseRNN} & 57.4 & 60.2 & 57.3 & 65.0 \\
        \bottomrule
    \end{tabularx}
    \label{tbl:lang-class}
\end{table}

To summarise, our probing experiments
suggest that the multilingual \system{CAE-RNN}'s embeddings encode more information about phonetic content, word duration, language identity, and speaker identity compared to the other multilingual models.

%% file: conclusion.tex
\section{Conclusion}

We proposed multilingual transfer as a way to obtain acoustic word embeddings for zero-resource languages.
We applied multilingual models trained jointly on six well-resourced languages to six zero-resource languages without any labelled data.
Three multilingual models---a Siamese recurrent neural network (RNN), a classifier RNN and a correspondence autoencoder (CAE) RNN with a reconstruction-like loss---consistently 
outperformed monolingual unsupervised models trained directly on the zero-resource language.
When fewer training languages are used, we showed that the multilingual CAE outperforms the other models and  that performance is affected by the combination of training and test languages.
These effects diminish as more training languages are used.
Since there was little to distinguish between the three multilingual models, we performed further analysis to determine the types of properties captured by the different models.
This revealed that the CAE is more sensitive than the other models to phonetic content, word duration information and language identity, but also (surprisingly) speaker identity.
Future work will look to apply these models in downstream systems and to consider more low-resource languages.

